\documentclass[a4paper, 12pt, times]{elsarticle}

\usepackage[utf8]{inputenc}
\usepackage{enumitem}
\usepackage{graphics}
\usepackage{amsmath}
\usepackage{amssymb}
\usepackage{subfig}
\usepackage{multirow}
\usepackage{algorithm}
\usepackage{algpseudocode}
\usepackage{setspace}
\usepackage{xcolor}
\usepackage{hyperref}
\usepackage{cleveref}
\usepackage{bm}
\usepackage{soul}
\usepackage{listings}
\usepackage{appendix}
\usepackage[framed,autolinebreaks,useliterate]{mcode}

\newcommand{\T}{\; \mathrm{T}}

\RequirePackage[margin=20mm]{geometry}

\graphicspath{{./figures/}}

\begin{document}

\begin{frontmatter}


\title{Machine Learning for Computational Science and Engineering – a brief introduction and some critical questions}

\author{Chennakesava Kadapa, \\
University of Bolton, Bolton BL3 5AB, United Kingdom. \\
Email: c.kadapa@bolton.ac.uk \\
\textbf{Version 1.0.0}
}

\begin{abstract}
Artificial Intelligence (AI) is now entering every sub-field of science, technology, engineering, arts, and management. Thanks to the hype and availability of research funds, it is being adapted in many fields without much thought. Computational Science and Engineering (CS\&E) is one such sub-field. By highlighting some critical questions around the issues and challenges in adapting Machine Learning (ML) for CS\&E, most of which are often overlooked in journal papers, this contribution hopes to offer some insights into the adaptation of ML for applications in CS\&E and related fields. This is a general-purpose article written for a general audience and researchers new to the fields of ML and/or CS\&E. This work focuses only on the forward problems in computational science and engineering. Some basic equations and MATLAB code are also provided to help the reader understand the basics.
\end{abstract}

\end{frontmatter}

\section{Introduction}
Artificial Intelligence (AI) is now everywhere; at least, that is how it appears to be. Every sector, every field uses AI in some or other form; at least, that is what they claim. AI is making its strides in every area, and Engineering is no exception. While AI has proven its worth for many applications, for example, pattern recognition, natural language processing, etc., its validity and worth in other fields, especially computational science and engineering (CS\&E), is yet to be established. This is not to say that AI is useless in computational science and engineering but to scrutinise its current use and highlight the issues and challenges.

When it comes to the use of AI in CS\&E, Machine Learning, particularly the algorithms based on artificial neural networks (ANN), matter the most. While it is nice to see fancy colourful contours and animations of results, the reality is not as colourful as it appears to be when it comes to the adaptation of ML for CS\&E. Several questions arise – some are open while many are ignored.
Looking at the top-level of adaption of ML for CS\&E, two questions are crucial.
\begin{itemize}
    \item Whether or not it is worth using ML in CS\&E?
    \item What are the requirements, bottlenecks, challenges and harms?
\end{itemize}

To be able to answer these two questions, or at least make a correct step in the right direction in this endeavour, one needs to understand, first of all, how CS\&E and ML work under the hood.

\subsection{Computational Science and Engineering (CS\&E)}
The forward problem in CS\&E involves (numerically) solving the partial differential equation (PDE) that govern the physics of the problem, e.g. solid mechanics, fluid mechanics, electromagnetism etc., for a given (i) geometry, (ii) material properties, (iii) constitutive laws and (iv) boundary and initial conditions. The solution is often obtained using a numerical technique, for example, finite difference method (FDM), finite volume method (FVM) or finite element method (FEM) etc. \cite{book-FEMFDM,book-Pinder}, implemented in the software of user's choice.

For the purpose of discussion, let us consider Poisson's equation in one dimension, which is one of the simplest differential equations, and it has numerous applications in heat transfer, electrostatics, diffusion, and solid mechanics. The one-dimensional Poisson equation for the variable of interest $y$ is given as
\begin{align} \label{eqn:poisson}
  - \, \frac{d^2y}{dx^2} = g, 
\end{align}
with the source term $g$ in the domain $x \in \left[ x_0, x_1 \right]$, and boundary conditions,
\begin{align}
  y(x=x_0) &= y_0, \mathrm{and} \\
  y(x=x_1) &= y_1.
\end{align}

\begin{figure}[H]
 \centering
 \includegraphics[trim=0mm 0mm 0mm 0mm, clip, scale=0.8]{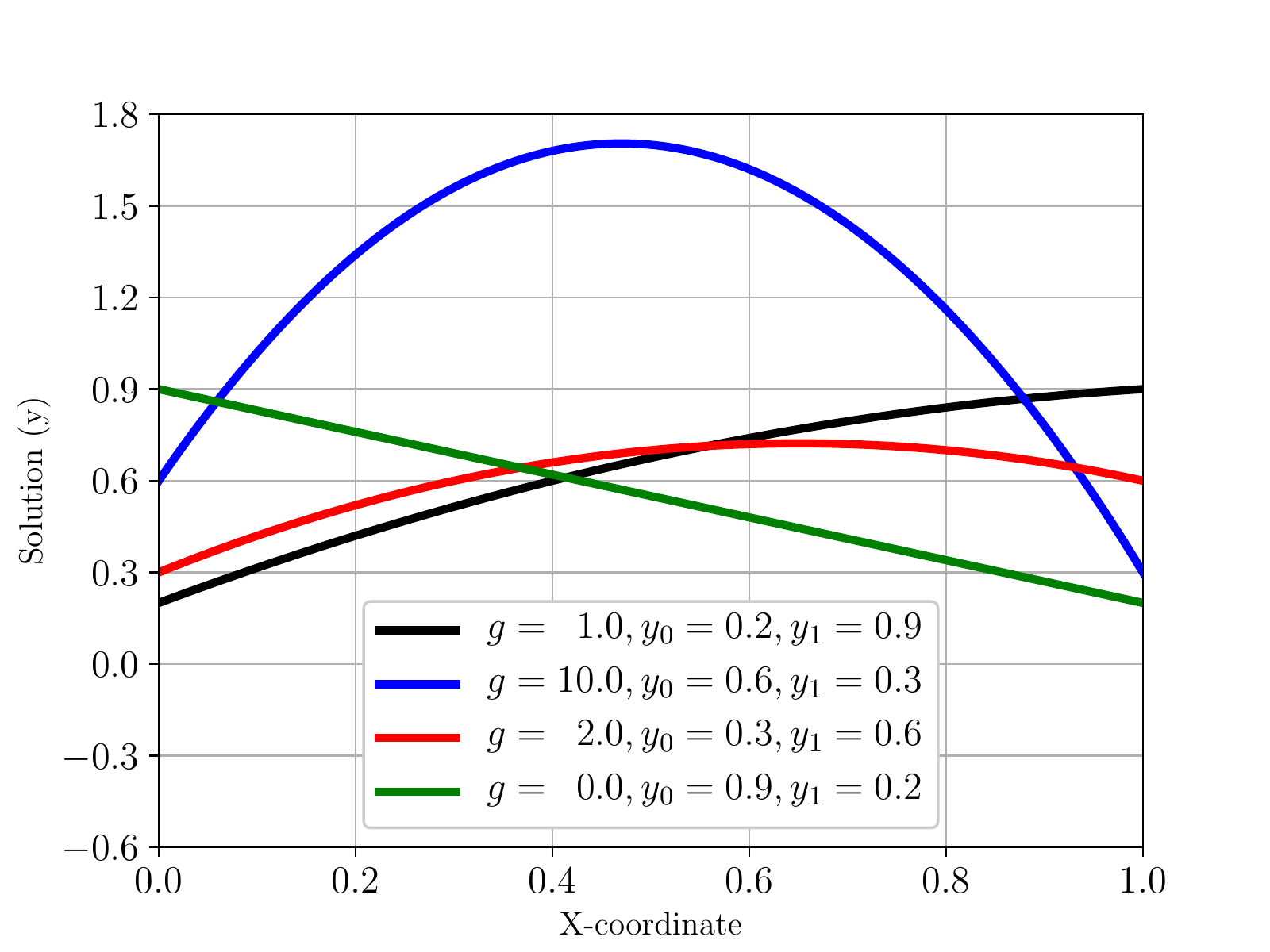}
 \caption{Solutions of the one-dimensional Poisson equation for a fixed domain $x \in \left[0,1 \right]$ for four different combinations of the source term $g$, and boundary conditions $y_0$ and $y_1$.}
 \label{fig:poisson}
\end{figure}

 Solutions of the one-dimensional Poisson equation (\ref{eqn:poisson}) for four different combinations of the source term $g$, and boundary conditions $y_0$ and $y_1$, are presented in Figure \ref{fig:poisson}. It is apparent that the solution is different for different values of the source term and boundary conditions. Therefore, even if we ignore the particular discretisation method used, or for that matter, how the software provides a solution for the given problem, we cannot and should not overlook that the solution of the PDE heavily depends on all the four inputs mentioned above. For the same PDE, the solution changes with a change in either of the four inputs: geometry, material properties, constitutive laws and boundary conditions. This necessitates that we must solve the PDE from scratch to account for the changes in either one or all four inputs of the problem, particularly so for any significant changes in shape and dimensions of the part and boundary conditions.

\subsection{Machine learning (ML)}
ML works completely different to CS\&E. There is input data and output data, and the ML algorithm finds the functional relation between the inputs and outputs. This functional relationship is found by \emph{fitting} the parameters using the \emph{least-squares method}. That is, ML based on ANN is nothing but (least-squares) fitting of input data in a hyper-dimensional space; at least, this is what is done in the usage of ML for CS\&E. There is only input and output data but no information about the underlying physics. In AI terminology, this type of learning is called \emph{supervised learning}, as the network is being \emph{supervised} using error measures based on the expected outputs during the training process. This can be understood relatively easily using \emph{linear regression} model \cite{book-Seber}.

\begin{figure}[H]
 \centering
 \includegraphics[trim=25mm 10mm 25mm 5mm, clip, scale=0.8]{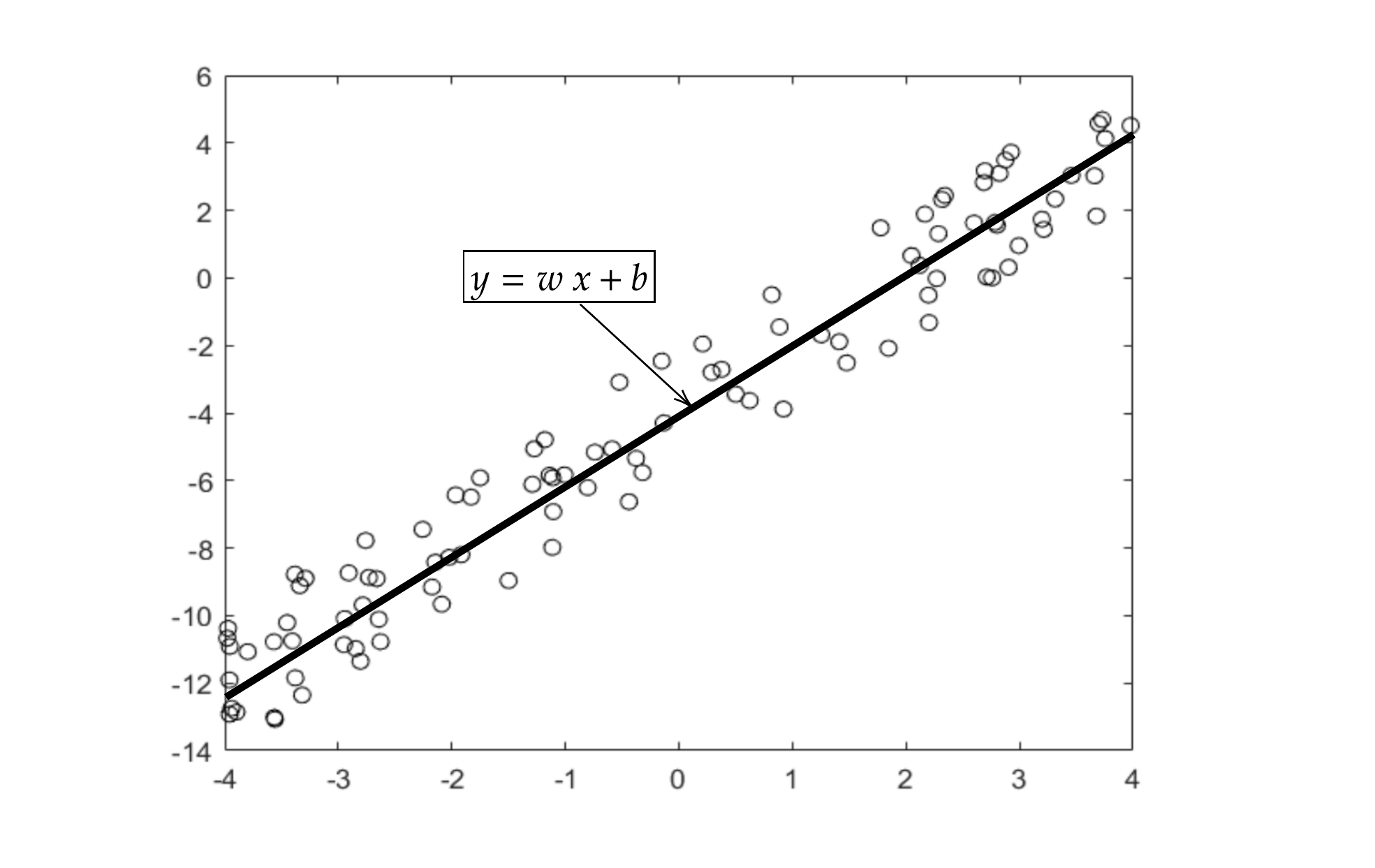}
 \caption{A typical example in linear regression. Here, $x$ is the input, $y$ is the output, $w$ is the slope, and $b$ is the intercept.}
 \label{fig:linearregr}
\end{figure}

Linear regression is an approach for modelling the relationship between the output and input, also known as the dependent and independent variables. A typical example in linear regression is shown in Figure \ref{fig:linearregr}. The objective is to find the slope ($w$) and intercept ($b$) in the linear relation $y=w x+b$. For the given set of data points, $w$ and $b$ are calculated based on some error measure, e.g. mean-squared error. This method is called the least-squares fitting. The corresponding equations and the associated MATLAB code for the least-squares fitting method are given in \ref{section-leastsquares}.

\begin{figure}[H]
 \centering
 \includegraphics[trim=25mm 10mm 25mm 10mm, clip, scale=1.0]{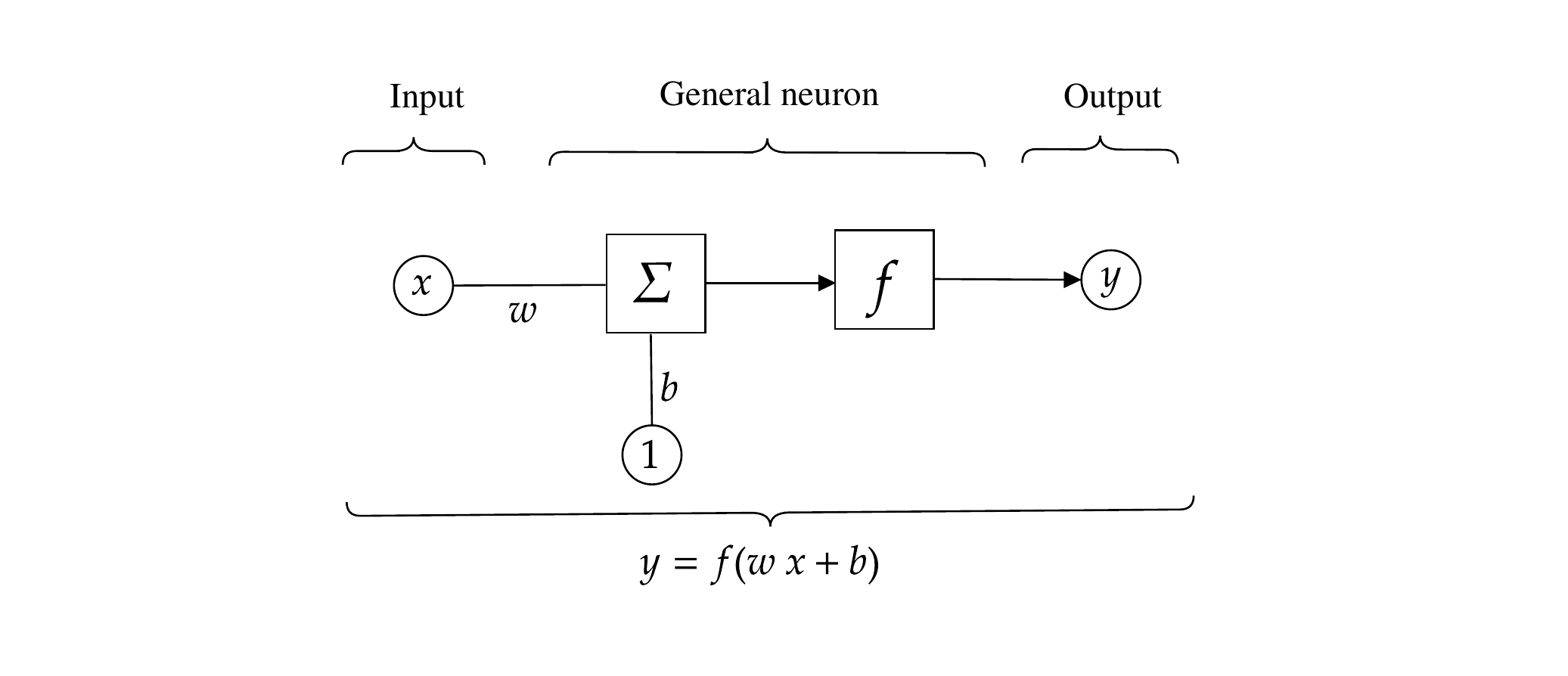}
 \caption{Linear regression model represented as a single-input single-output (SISO) neural network model. Here, $x$ is the input, $y$ is the output, $w$ is the weight, $b$ is the bias and $f$ is the transfer (or excitation) function.}
 \label{fig:nn-siso}
\end{figure}

Training of artificial neural networks works very similarly to least-squares fitting. A  representation of the linear regression model as a neural network model is shown in Figure \ref{fig:nn-siso}. In ML terminology \cite{book-Hagan}, the parameters $w$ and $b$ are called as weight and bias, respectively. The parameters $w$ and $b$ are calculated by minimising the \emph{loss function} using a numerical algorithm. This process of finding the weights and bias of a neural network model is called \emph{training}. It is only when a NN model is trained and tested thoroughly, it can be deployed for \emph{predictions}. Due to its applicability to complex, large-scale neural network models, the \emph{backpropagation algorithm} \cite{book-Hagan} is the widely used algorithm for calculating the weights and bias for typical applications in CS\&E. For the interested readers, the corresponding equations and the associated MATLAB code for the backpropagation algorithm used to train the neural network are given in \ref{section-backprop}. For the comprehensive details, refer to Hagan et al. \cite{book-Hagan} and Demuth et al. \cite{userguide-Demuth}. PyTorch \cite{paszke2019pytorch}, a Python-based software framework for machine learning, is an alternative to MATLAB's deep learning toolbox.

\begin{figure}[H]
 \centering
 \includegraphics[trim=25mm 0mm 25mm 10mm, clip, scale=0.9]{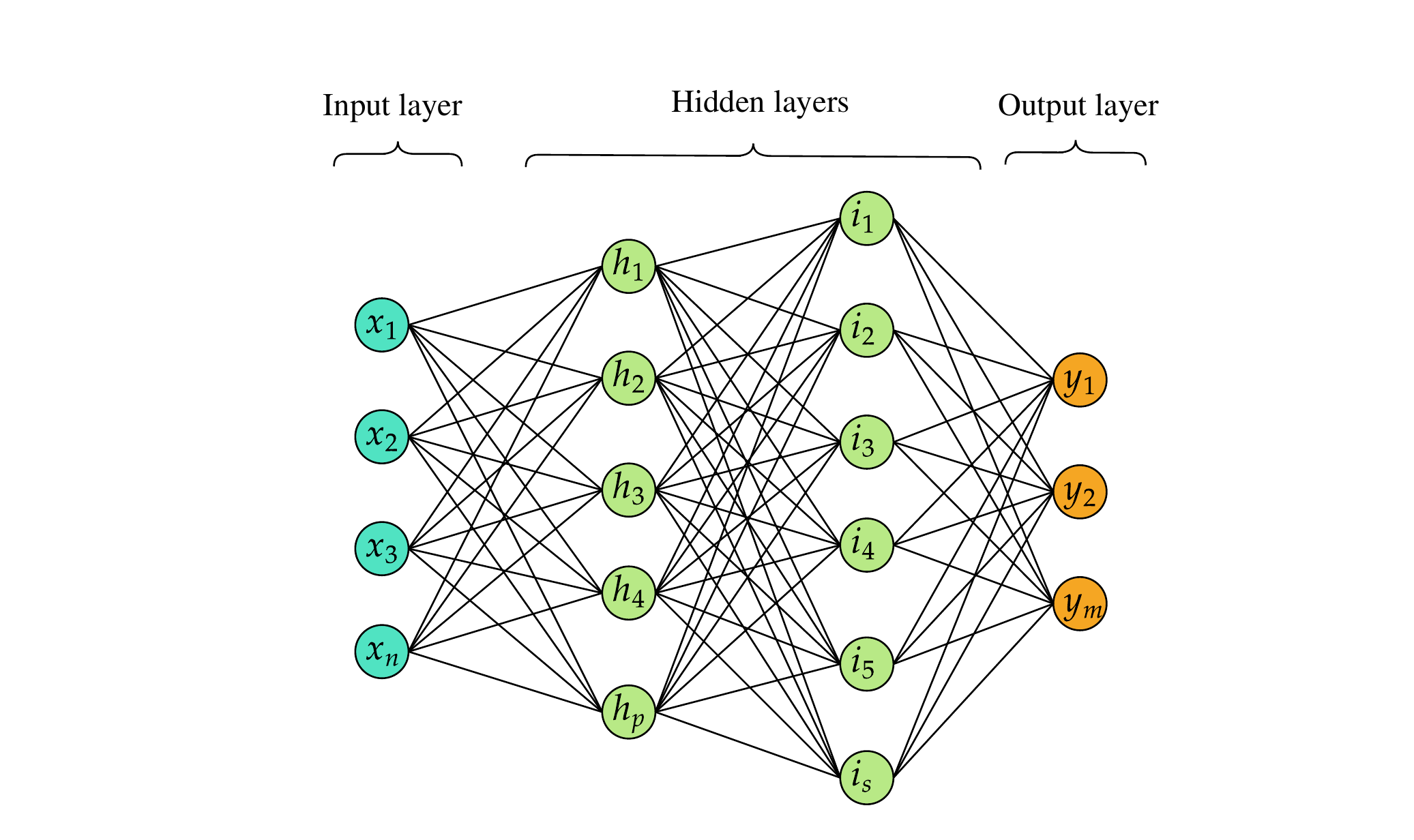}
 \caption{Multi-input multi-output (MIMO) neural network model.}
 \label{fig:nn-mimo}
\end{figure}

An extension of the neural network model to a more complex problem that consists of multiple inputs and multiple outputs is shown in Figure \ref{fig:nn-mimo}. The greater the number of inputs and outputs, the higher the number of hidden layers and the number of neurons in each hidden layer, which increases the number of unknown weight and bias parameters. On top of this, there is an additional requirement of choosing the excitation function for each (hidden) layer.

Therefore, ML based on ANN works completely different to the philosophy of CS\&E. A conventional ML algorithm knows nothing about the problem - neither the PDE nor any of the four inputs. It knows only about the input data and associated output data, popularly called the \emph{ground truth}. It must be emphasised here that \emph{ground truth} is not the exact solution; it is only a reference solution, often obtained using high-fidelity simulations. The recently introduced version of ANNs, namely Physics informed neural networks (PINN) \cite{RassiJCP2019}, or other variants of NN that claim to have accounted for physics, incorporate the PDE into the ML algorithm; this is done by replacing the conventional \emph{loss function} with the discrete version of the strong form of the associated PDE. Nevertheless, whether conventional NN or PINN, the fundamental idea is the same; the network gets trained based on the labelled data generated for a particular problem defined by the geometry, material properties and boundary conditions. PINN uses an improved loss function; however, it does not change, in any way, how ML works at the fundamental level. Besides, due to the inclusion of differential operators, one needs to use automatic differentiation techniques, which introduce additional numerical errors into the model, on top of the numerical errors coming from the simulation data and algorithms used for \emph{training}.

Once the network is trained sufficiently accurately, which requires a substantial amount of resources and significant effort from the user in collecting/preparing the data and going through the iterative process of training the NN model, the solution can be \emph{predicted} for \emph{a valid point} in the space of input parameters. It must be emphasised here that, although the trained network may compute the solution outside the physical domain by way of \emph{extrapolation}, such a solution outside the physical domain is meaningless. The only meaningful extrapolation would be for material and geometric properties beyond the range of the trained input data.

An ML model does not know anything about the PDE, geometry, boundary conditions or constitutive laws. To get an accurate prediction, a significant amount of data is required. The amount of data required for training is directly proportional to the number of input and output parameters and, therefore, the complexity of the associated neural network model required for that particular problem. You might be wondering, at this point is, \emph{"how will this data be generated?"}. Well, there is a magic tree on which this data grows. But, \emph{"how can one pick this data from the magic tree?"} is likely to be your next question. The answer is, by running hundreds, or perhaps of thousands or even more, of high-fidelity simulations. Remember, ML is data-hungry. This means that one must first carry out the high-fidelity simulations, thousands of them, to train a NN model.

\section{Key issues and questions}
It is not a question of whether ML works for applications in CS\&E; it does since it is all about \emph{fitting} the parameters. But, the main challenge for ML in CS\&E is not the ML technology itself but the data, particularly the process of generation of data that is used for training and testing of the NN model, and the computational resources required to accomplish this. Due to the user effort required to prepare hundreds of input files for different parameters, 3D problems compound the difficulties associated with data generation. We all can agree that most real-world engineering applications require solutions to 3D problems.

Moreover, even for the same PDE, the nitty-gritty details of a NN model: the number of layers, the number of neurons in each layer, the excitation function, the number of epochs etc., required to successfully train a NN within an acceptable level of accuracy vary from one problem to the other. This means that every major component that differs in geometry and shape requires a different NN model; when the geometry is changed significantly, a different NN model needs to be trained; when the initial and boundary conditions are changed considerably, the NN model needs to be re-trained, in the best-case scenario.

Therefore, it is not sufficient to show that the methodology works for simple 1D and/or 2D problems because it is established knowledge that one can \emph{fit} the curves to the data by tweaking functions and parameters. So, the conventional approach of demonstration for toy problems is useless when it comes to machine learning. The benefits of Machine Learning to CS\&E, if any, can only be understood from comprehensive studies of large-scale industrial 3D cases. Journal papers focused on 1D and 2D problems, for example, 1D advection-diffusion, 1D elastic wave propagation, 2D Navier-Stokes in square domains, are practically useless since a new NN model must be trained for each problem, irrespective of the dimension.

ML for applications in CS\&E may make sense only for problems that lack a functional relationship between output and input data, and such problems are not likely to be the ones defined by PDEs. For those problems where ML might be appropriate, due to the data-hungry nature of ML, its benefits over conventional methods are of primary concern.

An important aspect that is often overlooked is the cost of generation and preparation of data and training. Most of the papers on ML for CS\&E present only equations and colour contour plots comparing the predicted solution with some reference solution (aka ground truth). There are several crucial details of the ML models that these papers omit, for example, number of layers, number of neurons, excitation function, tolerances, computing resources, etc. This omission is believed to be due to some obvious reasons. However, these details are crucial for any user training a NN model. Some important questions on this aspect are:
\begin{enumerate}[label=Q\arabic*)]
  \item	What are the computational resources used?
  \item What is the cost of generating the input data?
  \item What is the cost of training a neural network?
  \item What are the number of layers, number of neurons in each layer, and the excitation function for each layer, and how to choose them?
  \item How to avoid over-fitting and under-fitting?
  \item What are the tolerances and learning rate used and how they affect the training, for example, number of epochs?
  \item How much data is necessary to obtain a reasonably accurate solution to a problem?
\end{enumerate}

The above-listed questions are not the ones that are actually concerned with the fundamental aspects of the core problems in CS\&E. Nevertheless, their importance concerns the development of a good NN model. This is what makes ML a challenging technology to make any meaningful impact in CS\&E. All of these questions are concerned with how ML operates and are problem-specific even in a given field. Moreover, questions particularly concerning data - its collection and preparation, and training of NN model, are crucial to the success of any project on the use of ML in CS\&E. Without discussing the details of resources required for collecting/preparing data and training the NN model, which is substantial for 3D problems, the adaption of ML for practical engineering problems is going to be more of a pipe dream than reality. One or two start-ups and established firms might have done this, but at what cost?

Moreover, concerned with the prediction of a trained NN model, the following three additional questions arise:
\begin{enumerate}[label=Q\arabic*)]
  \setcounter{enumi}{7}
  \item How accurate is the output of the trained network beyond the range of material parameters considered in the input?
  \item How does the trained network generalise to other problems, e.g. changes in geometry, topology, discretisation, initial conditions, etc.?
  \item What are the margins of error and the respective tolerances for input parameters? For example, how much can the input parameters vary for a given error margin?
\end{enumerate}

Most of the published papers on the use of ML for CS\&E do not discuss anything on questions Q8, Q9 and Q10. But these are essential if such NN models were to be deployed as surrogate models in digital twin technologies.

Furthermore, it is not fair to compare the prediction time of a trained NN model against that of high-fidelity simulations. Since the data for training and testing of NN models must be generated by running high-fidelity simulations of hundreds or thousands of cases, which requires a substantial amount of computational resources, this cost cannot be ignored. Therefore, the total time ($T_t$) for predicting the end result using a NN model is the sum of time taken for data generation ($T_{dg}$), time taken for network training ($T_{nt}$) and time taken for predictions ($T_{pr}$). That is,
\begin{align}
    T_t = T_{dg} + T_{nt} + N \, T_{pr}.
\end{align}
with $N$ as the number of predictions in the deployment stage.

Taking all the costs into consideration, the use of ML for CS\&E would prove worthwhile only when the number of predictions in the deployment stage is so substantial that the time for predictions overshadows the cost of data generation and training beyond a break-even point.

\section{Summary and conclusion}
This contribution provides a brief introduction to the application of Machine Learning in computational science and engineering. A set of ten questions, which are often overlooked in journal papers on ML for CS\&E yet essential to practitioners, are highlighted in the hope that they help the novice users and practitioners to better their understanding of the adaptation of ML for CS\&E. 

A lack of an honest discussion around these questions is to be treated as a cause for concern because the points highlighted are important for the end-users gives the non-experts and researchers from other fields a false impression of the potential of ML for CS\&E. This can be particularly dangerous when ML is adapted for applications in fields such as biomedical and nuclear engineering. Can doctors use ML for cardiovascular flow simulations? Since cardiovascular simulations are patient-specific, is it possible to obtain a sufficient number of data samples to train a NN model to compute results that the doctors can trust? Can we use ML for applications in nuclear engineering, where the price we must pay for a miscalculation is quite heavy? These are still open questions.

Without answering all these questions and addressing the associated issues, the adaptation of ML for real-world applications of computational science and engineering might prove to be a costly and risky affair. The onus is on the CS\&E community, particularly those adopting ML for CS\&E. Academia must follow ethical engineering practices and facilitate honest discussions that can help solve our problems rather than innovating for the namesake of fantasising and sensationalising technologies, which is counter-productive.


\begin{appendices}


\renewcommand{\theequation}{A.\arabic{equation}}
\renewcommand{\thesection}{Appendix A}

\setcounter{equation}{0}

\newpage
\section{Linear regression and least-squares fitting} \label{section-leastsquares}
The functional relation in linear regression is model given by
\begin{align} \label{eqn-linearreg-2}
 y = w x + b.
\end{align}

If there are $N$ data points $(x_1,y_1)$, $(x_1,y_1)$, $\ldots$,  $(x_N,y_N)$, we can write equation (\ref{eqn-linearreg-2}) in the matrix-vector format as
\begin{equation}
\mathbf{y}
\;
=
\;
\begin{bmatrix}
y_{1}\\
y_{2}\\
\vdots \\
y_{N}
\end{bmatrix}
\;
=
\;
\begin{bmatrix}
x_{1} & 1 \\
x_{2} & 1 \\
\vdots  & \vdots \\
x_{N} & 1
\end{bmatrix}
\;
\begin{bmatrix}
w \\
b
\end{bmatrix}
=
\;
\mathbf{X} \, \boldsymbol{\beta}
\quad
\mathrm{or}
\quad
\mathbf{y}
\;
=
\;
\begin{bmatrix}
x_{1} \\
x_{2} \\
\vdots \\
x_{N}
\end{bmatrix}
w
+
\begin{bmatrix}
1 \\
1 \\
\vdots \\
1
\end{bmatrix}
b
=
\mathbf{x} \, w + \mathbf{1} \, b,
\end{equation}
from which the error vector becomes,
\begin{align}
    \mathbf{e} = \mathbf{y} - \mathbf{X} \, \boldsymbol{\beta}
\end{align}

The unknowns $w$ and $b$ are computed by minimising the error (or loss) function. Sum of squared errors is the widely used loss function, and it is given as
\begin{equation}
L
= \mathbf{\| y} -\mathbf{X} \, \boldsymbol{\beta} \| ^{2}
= \left[\mathbf{y} -\mathbf{X} \, \boldsymbol{\beta}\right]^{\T} \left[\mathbf{y} -\mathbf{X} \, \boldsymbol{\beta}\right]
= \mathbf{e}^{\T} \, \mathbf{e}.
\end{equation}

For $\displaystyle L$ to be minimum with respect to $\displaystyle \boldsymbol{\beta }$, its derivative should vanish, i.e.,
\begin{equation}
\frac{\partial L}{\partial \boldsymbol{\beta}} = 0,
\end{equation}
which leads to the equation for unknowns ($\boldsymbol{\beta}$) as
\begin{equation}
\boldsymbol{\beta} = \mathbf{X}^{+} \, \mathbf{y}, \quad \mathrm{with} \quad \mathbf{X}^{+} = \left( \mathbf{X}^{\T} \, \mathbf{X}\right)^{-1} \mathbf{X}^{\T},
\end{equation}
where $\mathbf{X}^{+}$ is the Moore-Penrose inverse, also called the pseudoinverse.

In the literature, this procedure is called \emph{the least-squares fitting}. The MATLAB code for calculating the parameters using the least-squares fitting is given in code block \ref{listing-regression}.

\newpage
\begin{lstlisting}[language=MATLAB,caption={MATLAB code using the least-squares fitting method.},captionpos=b, label=listing-regression]
clear all; clc;
N = 100; % number of points

w =  2.0;    b = -4.0;
x_true = linspace(-4.0,4.0,N);
y_true = w*x_true + b;

% generate data
rng(1,'philox'); % set the seed of random number generator

x = -4.0 + (4.0-(-4.0))*rand(N,1);
e = -2.0 + (2.0-(-2.0))*rand(N,1);    % noise
y = w*x + b + e;

X = [x ones(N,1)];    % matrix X

params = pinv(X)*y;   % solver for parameters

w_hat = params(1)
b_hat = params(2)

% Plotting
plot(x, y, 'ko');
hold on;
plot(x_true, y_true, 'r', linewidth=2);
hold on;
y_fit = w_hat*x_true + b_hat;
plot(x_true, y_fit, 'k', linewidth=2);
legend('Data', 'True model', 'Fitted model', 'Location', 'southeast');
\end{lstlisting}

\renewcommand{\theequation}{B.\arabic{equation}}
\renewcommand{\thesection}{Appendix B}

\setcounter{equation}{0}

\newpage
\section{Artificial Neural Networks and backpropagation algorithm} \label{section-backprop}
The functional relation between the output and input in an artificial neural network model shown in Figure \ref{fig:nn-siso} is given by
\begin{align}
    y = f(n), \; \mathrm{with} \; n = w \, x + b,
\end{align}
where $n$ is the output of the neuron and $f$ is the transfer (or excitation) function.

Accordingly, the data points can be written in the matrix-vector form as
\begin{align}
    \mathbf{y} = f(\mathbf{X} \, \boldsymbol{\beta}),
\end{align}
for which the error vector becomes,
\begin{align}
    \mathbf{e} = \mathbf{y} - f(\mathbf{X} \, \boldsymbol{\beta}).
\end{align}

The loss function is assumed to be the sum of squared errors. That is,
\begin{equation}
L
= \mathbf{\| y} - f(\mathbf{X} \, \boldsymbol{\beta}) \| ^{2}
= \left[\mathbf{y} - f(\mathbf{X} \, \boldsymbol{\beta}) \right]^{\T} \left[\mathbf{y} - f(\mathbf{X} \, \boldsymbol{\beta}) \right] = 
\mathbf{e}^{\T} \, \mathbf{e}.
\end{equation}

The parameters $w$ and $b$ are determined by minimising the loss function, similar to the least-squares method. However, instead of solving directly as in the least-squares method, they are calculated using the backpropagation algorithm \cite{book-Hagan} in which the parameters are updated iteratively until the chosen convergence criterion is satisfied. In the backpropagation algorithm, the error computed at the end of the feed-forward (or forward propagation) step is \emph{propagated} backwards to update the weights and bias; hence, the name. Since the weights are \emph{updated} iteratively, the algorithm starts with an initial guess for each parameter.

Different methods, for example, steepest descent, Newton's, Gauss-Newton's and Levenberg-Marquardt, are used to update the weights \cite{book-Hagan, book-Nocedal, book-Bjorck}. The Levenberg-Marquardt method is commonly used for training small- and medium-sized networks, but it is mathematically more complicated compared to the rest. For demonstration, the parameters are updated using the steepest descent algorithm.

Following the steepest descent algorithm, we get
\begin{align}
w_{( k+1)} =w_{( k)} \ -\alpha \ \frac{\partial L}{\partial w},
\end{align}

\begin{align}
b_{( k+1)} =b_{( k)} \ -\alpha \ \frac{\partial L}{\partial b},
\end{align}
where $\displaystyle \alpha $ is the training rate and $\displaystyle k$ is the iteration counter. From the loss function, we get
\begin{align}
\frac{\partial L}{\partial w} = 2 \, \mathbf{e}^{\T} \, \frac{\partial \mathbf{e}}{\partial w} = - \, 2 \, \mathbf{e}^{\T} \, \frac{\partial f}{\partial n} \, \frac{\partial (\mathbf{X} \, \boldsymbol{\beta})}{\partial w} = - \, 2 \, \mathbf{e}^{\T} \, \frac{\partial f}{\partial n} \, \mathbf{x}, 
\end{align}

\begin{align}
\frac{\partial L}{\partial b} = 2 \, \mathbf{e}^{\T} \, \frac{\partial \mathbf{e}}{\partial b} = - \, 2 \, \mathbf{e}^{\T} \, \frac{\partial f}{\partial n} \, \frac{\partial (\mathbf{X} \, \boldsymbol{\beta})}{\partial b} = - \, 2 \, \mathbf{e}^{\T} \, \frac{\partial f}{\partial n} \, \mathbf{1}.
\end{align}

Several different transfer functions are available that are appropriate to the application at hand \cite{book-Hagan}. If the transfer function is assumed to be a purely linear transfer function, we have
\begin{align}
    f(n) = n, \quad \mathrm{and} \quad \frac{\partial f}{\partial n} = 1.
\end{align}

The backpropagation algorithm is sensitive not only to the initial guesses of the weight and bias parameters but also to the learning rate parameter $\alpha$. For guaranteed convergence, the learning rate ($\alpha$ should be small enough; however, for smaller values of $alpha$, the convergence of iterations is slower. The interested user can experiment with different values of $alpha$ and initial guess for $w$ and $b$ using the MATLAB code for the backpropagation given in code block \ref{listing-backprop}.

The backpropagation algorithm, in fact, the entire training process, is available as a few functions in MATLAB's Deep Learning Toolbox or any other sophisticated software libraries such as TensorFlow. A sample MATLAB code using MATLAB's Deep learning toolbox for the problem of linear regression is presented in code block \ref{listing-DLT}.

\newpage
\begin{lstlisting}[caption={MATLAB code using backpropagation algorithm.},captionpos=b, label=listing-backprop]
clear all; clc;
N = 100; % number of points

w =  2.0;    b = -4.0;
x_true = linspace(-4.0,4.0,N);
y_true = w*x_true + b;

% generate data
rng(1,'philox'); % set the seed of random number generator

x = -4.0 + (4.0-(-4.0))*rand(N,1);
e = -2.0 + (2.0-(-2.0))*rand(N,1);    % noise
y = w*x + b + e;

w_hat = 1.0;   b_hat = -1.0;   % initial guess
alpha = 0.001;                 % learning rate

loss_fun = zeros(1,N);
loss = 1.0e-10;
for iter=1:N
    err = y - (w_hat*x + b_hat);
    
    lossPrev = loss;
    loss     = sum(err.^2);
    loss_fun(iter) = loss;

    if( abs(loss-lossPrev) < 1.0e-6)
        break;
    end
    w_hat = w_hat + alpha*2.0*err'*x;
    b_hat = b_hat + alpha*2.0*sum(err);
end
plot(x, y, 'ko');    hold on;
plot(x_true, y_true, 'r', linewidth=2);    hold on;
y_fit = w_hat*x_true + b;
plot(x_true, y_fit, 'k', linewidth=2);
legend('Data', 'True model', 'Fitted model', 'Location', 'southeast');
figure();    plot(log10(loss_fun))
\end{lstlisting}

\begin{lstlisting}[caption={MATLAB code using MATLAB's Deep Learning Toolbox.},captionpos=b, label=listing-DLT]
clear all; clc;
N = 100; % number of points

w =  2.0;    b = -4.0;
x_true = linspace(-4.0,4.0,N);
y_true = w*x_true + b;

% generate data
rng(1,'philox'); % set the seed of random number generator

x = -4.0 + (4.0-(-4.0))*rand(N,1);
e = -2.0 + (2.0-(-2.0))*rand(N,1);    % noise
y = w*x + b + e;

% Set the neural network model
hiddenLayerSize = 1;    % One neuron in the hidden layer
trainFcn = 'trainlm';   % Levenberg-Marquardt algorithm

net = fitnet(hiddenLayerSize, trainFcn); % Fitting neural network

net.layers{1}.transferFcn = 'purelin'; % Pure linear transfer function

% Division of data for Training, Validation, Testing
net.divideParam.trainRatio = 80/100;  % 80% for training
net.divideParam.valRatio   = 10/100;  % 10% for validation
net.divideParam.testRatio  = 10/100;  % 10% for testing

net = train(net, x', y'); % Train the network
y_fit = net(x_true);      % Test the network

% Plotting
plot(x, y, 'ko');
hold on;
plot(x_true, y_true, 'r', linewidth=2);
hold on;
plot(x_true, y_fit, 'k', linewidth=2);
legend('Data', 'True model', 'Fitted model', 'Location', 'southeast');
\end{lstlisting}

\end{appendices}


\end{document}